\documentclass[final]{cvpr}

\usepackage{times}
\usepackage{epsfig}
\usepackage{graphicx}
\usepackage{amsmath}
\usepackage{amssymb}

\usepackage{mathtools}
\usepackage[dvipsnames]{xcolor}
\usepackage{pifont}
\newcommand{\cmark}{{\color{Green}\ding{51}}}%
\newcommand{\xmark}{{\color{Red}\ding{55}}}%
\usepackage{hhline}
\usepackage{multirow}
\usepackage{bm}
\usepackage{placeins} 
\usepackage{flafter}  
\usepackage{balance}
\usepackage{lipsum}
\usepackage{enumitem}
\setlist[itemize]{noitemsep, topsep=3pt}
\setlist[enumerate]{noitemsep, topsep=3pt}

\usepackage[linesnumbered,ruled,vlined]{algorithm2e}

\SetCommentSty{mycommfont}
\SetKwInput{KwInput}{Input}                
\SetKwInput{KwOutput}{Output}     
\SetKw{Continue}{continue}

\usepackage{setspace}

\DeclareMathOperator*{\argmin}{\arg\!\min}

\newlength{\textfloatsepsave} 
\newlength{\floatsepsave} 

\newdimen\origiwspc
\newdimen\origiwstr

\usepackage{array}
\makeatletter
\newcommand{\thickhline}{%
    \noalign {\ifnum 0=`}\fi \hrule height 1pt
    \futurelet \reserved@a \@xhline
}
\newcolumntype{"}{@{\hskip\tabcolsep\vrule width 1pt\hskip\tabcolsep}}
\makeatother

\usepackage[pagebackref=true,breaklinks=true,colorlinks,bookmarks=false]{hyperref}

\def\cvprPaperID{2029} 
\def\confYear{CVPR 2021}

\begin{document}

\title{Rotation-Only Bundle Adjustment}

\author{Seong Hun Lee\thanks{This work was partially supported by the Spanish govt. (PGC2018- 096367-B-I00) and the Arag{\'{o}}n regional govt. (DGA-T45{\_}17R/FSE).} \hspace{25pt} Javier Civera \\
I3A, University of Zaragoza, Spain\\
{\tt\small \{seonghunlee, jcivera\}@unizar.es}
}

\maketitle

\begin{abstract}
   We propose a novel method for estimating the global rotations of the cameras independently of their positions and the scene structure.
   When two calibrated cameras observe five or more of the same points, their relative rotation can be recovered independently of the translation. 
   We extend this idea to multiple views, thereby decoupling the rotation estimation from the translation and structure estimation.
   Our approach provides several benefits such as complete immunity to inaccurate translations and structure, and the accuracy improvement when used with rotation averaging.
   We perform extensive evaluations on both synthetic and real datasets, demonstrating consistent and significant gains in accuracy when used with the state-of-the-art rotation averaging method.
\end{abstract}

\vspace{-1em}
\section{Introduction}
Bundle adjustment is the problem of reconstructing the camera poses (\ie, rotations and translations) and the 3D scene structure from the image measurements.
It plays a crucial role in many areas of 3D vision, such as structure from motion \cite{hartley_book}, visual odometry \cite{scaramuzza2011visual}, and simultaneous localization and mapping \cite{cadena2016past}.  
For this reason, significant research endeavors have been devoted to this problem, which led to tremendous progress over the past two decades.

Bundle adjustment aims to obtain jointly optimal structure and camera poses by minimizing the image reprojection errors \cite{triggs2000bundle}.
Being a nonlinear optimization problem, it requires a good initialization to ensure the convergence to the statistically optimal solution \cite{hartley_book}.
A common strategy involves the following steps:
(1) Estimate the pairwise motions.
(2) Estimate the global rotations through rotation averaging (\eg, \cite{arrigoni2018robust, chatterjee2018robust}).
(3) Estimate the global translations (\eg, \cite{govindu2001combining, wilson20141dsfm}).
(4) Triangulate the points (\eg, \cite{kang2014robust, our_multiview_triangulation}).

In such a pipeline, it is important to make an accurate initial guess of the rotations, as the subsequent steps directly depend on it.   
To this end, one could try to improve the rotation averaging method or its input (\ie, the relative pairwise motion estimates).
Recent examples of the former include \cite{arrigoni2018robust, chatterjee2018robust,  dellaert2020shonan, anders2019rotation, purkait2020neurora} and the latter include \cite{brachmann2019neural, briales2018certifiably, garcia2020certifiable, zhao2019efficient}.

These two types of approaches are certainly useful for initializing the rotations.
However, relative pose estimation is limited to two views only, while rotation averaging does not directly leverage the image measurements.
That is, it treats all relative rotations equally even if they were estimated from different numbers of points with different noise statistics and distributions.
To our knowledge, no previous work has addressed this limitation for rotation estimation.

In this work, we present a novel method that, given the initial estimates of the rotations, performs rotation-only optimization using the image measurements as direct input.
Our work is based on \cite{kneip_rotation_ICCV}, where it was proposed to optimize the rotation between two views independently of the translation.  
We extend this idea to multiple views.
We call our approach \textit{rotation-only bundle adjustment} because it can be seen as the decoupling of the rotation estimation from the translation and structure estimation in bundle adjustment.
This provides the following advantages:
\begin{itemize}\itemsep0em
    \item The rotations are estimated without requiring the knowledge of the translations and structure.
    This greatly simplifies the optimization problem.    
    \item The rotations are immune to inaccurate estimation of the translations and structure.
    \item Both pure and non-pure rotations are treated in a unified manner, as we do not need to triangulate and discard the low-parallax points.
    \item It can be used after rotation averaging to improve the accuracy of the rotation estimates.
\end{itemize}
Table \ref{tab:method_comparison} summarizes the differences between our method and the related methods.

\begin{table*}[t]
\vspace{-0.1em}
\begin{center}
\begin{tabular}{c|c|c|c|}
\cline{2-4}
 & Independent of the translations & Directly using the image & Applicable \\
 & and the 3D scene structure? & measurements as input? & to $n$ views?\\
\hline
\multicolumn{1}{|c|}{Full bundle adjustment (\eg, \cite{triggs2000bundle})} & \xmark & \cmark & \cmark \\
\multicolumn{1}{|c|}{Rotation averaging (\eg, \cite{chatterjee2018robust})} & \cmark & \xmark & \cmark\\
\multicolumn{1}{|c|}{Direct rotation optimization \cite{kneip_rotation_ICCV}} & \cmark & \cmark & \xmark\\
\multicolumn{1}{|c|}{Rotation-only bundle adjustment} & \cmark & \cmark &\cmark\\
\hline
\end{tabular}
\vspace{-0.5em}
\end{center}
\caption{
Comparison between the related methods. 
To the best of our knowledge, we are the first to propose a multiview rotation-only optimization method using the image measurements as direct input.
Our method can be generalized to both pure and non-pure rotations.
}
\label{tab:method_comparison}
\vspace{-0.5em}
\end{table*}


The paper is organized as follows:
In the next two sections, we review the related work and the preliminaries.
Section \ref{sec:review_kneip} reviews the two-view rotation-only method by Kneip and Lynen \cite{kneip_rotation_ICCV}.
We describe our method in Section \ref{sec:method} and show the experimental results in Section \ref{sec:results}.
Finally, Section \ref{sec:discussions} and \ref{sec:conclusion} present discussions and conclusions.

To download our code and the supplementary material, go to \url{https://seonghun-lee.github.io}.

\section{Related Work}
\label{sec:related}
\vspace{-0.3em}
Our work is related to several areas of study in 3D vision and robotics, namely structure from motion, simultaneous localization and mapping, bundle adjustment, rotation averaging, and relative pose estimation.  

\textbf{Structure from motion (SfM)} is the problem of recovering the camera poses and the 3D scene from an unordered set of images. 
Large-scale systems may handle from hundreds of thousands \cite{building_rome} to millions of images \cite{reconstructing_world, zhu2018very}.
We refer to \cite{survey_sfm, sfm_revisited} for excellent reviews of the SfM literature.
The backbone of most SfM systems is bundle adjustment, the joint optimization of the camera poses and the 3D structure.
To obtain the optimal solution, it requires good initial estimates of the poses and points \cite{hartley_book}.
In many works \cite{arie2012global, cui2015linear, cui2015global, enqvist2011non, govindu2001combining, martinec2007robust, moulon2013global, wilson20141dsfm, ozyesil2015robust}, the initialization consists of the following steps:
(1) Estimate the relative poses between the camera pairs observing many points in common.
(2) Perform multiple rotation averaging.
(3) Estimate the camera locations (and the 3D points).
In such a pipeline, one may use our method as Step 2.5 to refine the rotations.

\textbf{Simultaneous localization and mapping (SLAM)} is the problem of estimating the camera motion and the 3D scene in real time.
Like SfM, the modern SLAM systems rely on bundle adjustment to jointly optimize the keyframe poses and the map points \cite{orb_slam3, dso, lcsd_slam, orb_slam}.
Recently, in \cite{chng2020monocular, why_bundle_adjust}, it was suggested that decoupling the rotation estimation through rotation averaging improves the efficiency and the handling of pure rotations.

\textbf{Bundle adjustment} is mainly classified into geometric and photometric methods.
The former minimizes the reprojection errors \cite{agarwal2010bundle, hartley_book, sba, triggs2000bundle}, and the latter minimizes the photometric errors \cite{alismail2017photometric, delaunoy2014photometric, dso, woodford2015large}.
The bundle adjustment problems have been studied extensively for several decades, which led to diverse techniques for improving the scalability (\eg, \cite{agarwal2010bundle, ssba, g2o, sba}) and the accuracy (\eg, \cite{triggs2000bundle, zach2014robust, zach2018descending}).
In \cite{hong2018pose}, an initialization-free approach was proposed.
To our knowledge, however, no previous work has completely decoupled the rotation estimation in bundle adjustment.

\textbf{Rotation averaging} takes two forms: single rotation averaging that averages several estimates of a single rotation to obtain the best estimate \cite{hartley2011L1, robust_single_rotation_averaging}, and multiple rotation averaging that finds the multiple rotations $\mathbf{R}_i$ given several noisy constraints on the relative rotations $\mathbf{R}_i\mathbf{R}_j^\top$ \cite{arie2012global, arrigoni2018robust,   chatterjee2018robust,dellaert2020shonan, anders2019rotation,  hartley2011L1, martinec2007robust}.
We refer to \cite{hartley_rotation_averaging, tron2016survey} for an excellent tutorial and survey on the topic.
As discussed earlier, multiple rotation averaging has wide application to SfM.
This problem differs from bundle adjustment in that (1) only rotations are estimated, and (2) the input is the relative rotation estimates, not the image measurements. 
Since the states and the input are small compared to bundle adjustment, the computation process is faster and simpler.
However, the downside is that it does not directly reflect the errors with respect to the image measurements.
In contrast, our optimization method directly uses the image measurements as input, while maintaining only rotations in the state space.
In this context, our method can be seen as the middle ground between rotation averaging and bundle adjustment.

\textbf{Relative pose estimation}
Given a set of five or more point matches between two calibrated views,  their relative pose can be obtained using a minimal method (\eg, \cite{quest, li2006five, kukelova2008polynomial, nister_five_point_PAMI}) with RANSAC \cite{brachmann2019neural,ransac,usac} or a non-minimal method (\eg,  \cite{briales2018certifiably, in_defense_of_8point, kneip_rotation_ICCV, zhao2019efficient}).
In \cite{kneip_rotation_ICCV,kneip2012finding}, it was shown that the rotation can be estimated independently of the translation.
In this work, we extend the idea of \cite{kneip_rotation_ICCV} to $n\geq2$ views by aggregating multiple two-view costs and minimizing it through iterative nonlinear optimization.

\section{Preliminaries and Notation}
\label{sec:prelim}
\vspace{-0.3em}
We use bold lowercase letters for vectors, bold uppercase letters for matrices, and light letters for scalars.
We denote the Hadamard product, division and square root by ${\mathbf{A}\circ\mathbf{B}}$, ${\mathbf{A}\oslash\mathbf{B}}$ and ${\mathbf{A}^{\circ1/2}}$, respectively.
For a 3D vector $\mathbf{v}$, we define $\mathbf{v}^\wedge$ as the corresponding $3\times3$ skew-symmetric matrix, and denote the inverse operator by $(\cdot)^\vee$, i.e., $\left(\mathbf{v}^\wedge\right)^\vee=\mathbf{v}$.
The Euclidean norm of $\mathbf{v}$ is denoted by $\lVert\mathbf{v}\rVert$, and its unit vector by $\widehat{\mathbf{v}}=\mathbf{v}/\lVert\mathbf{v}\rVert$.
A rotation matrix $\mathbf{R}\in SO(3)$ can be represented by the corresponding rotation vector $\mathbf{u}=\theta\widehat{\mathbf{u}}$, where $\theta$ and $\widehat{\mathbf{u}}$ represent the angle and the unit axis of the rotation, respectively.
The two representations are related by Rodrigues formula, and we denote the mapping between them by $\text{Exp}(\cdot)$ and $\text{Log}(\cdot)$ \cite{forster2017onmanifold}:
\begin{equation}
    \mathbf{R}=\mathrm{Exp}(\mathbf{u})
    :=
    \mathbf{I}+\frac{\sin\left(\lVert\mathbf{u}\rVert\right)}{\lVert\mathbf{u}\rVert}\mathbf{u}^\wedge+\frac{1-\cos\left(\lVert\mathbf{u}\rVert\right)}{\lVert\mathbf{u}\rVert^2}\left(\mathbf{u}^\wedge\right)^2,\vspace{-0.5em}
\end{equation}
\begin{equation}
\label{eq:log_map1}
    \mathbf{u}=\mathrm{Log}(\mathbf{R}):=
    \frac{\theta}{2\sin(\theta)}\left(\mathbf{R}-\mathbf{R}^\top\right)^\vee 
\end{equation}
\begin{equation}
\label{eq:log_map2}
    \text{with} \quad \theta = \arccos\left(\left(\mathrm{tr}(\mathbf{R})-1\right)/2\right).
\end{equation}
We denote the 3D position of a point with index $i$ in a world reference frame $w$ as $(\mathbf{x}_i)_w=[(x_i)_w, (y_i)_w, (z_i)_w]^\top$ and a perspective camera with index $j$ as $c_j$.
In the reference frame of $c_j$, the position of $\mathbf{x}_i$ is given by $(\mathbf{x}_i)_j=[(x_i)_j, (y_i)_j, (z_i)_j]^\top=\mathbf{R}_{j}(\mathbf{x}_i)_w+\mathbf{t}_{j}$,
where $\mathbf{R}_{j}$ and $\mathbf{t}_{j}$ are the rotation and translation that relate the local reference frame of $c_j$ to the world.
The projection of $\mathbf{x}_i$ in the image plane of $c_j$ has the pixel coordinates $[(u_i)_j, (v_i)_j]^\top=\begin{bsmallmatrix}1&0&0\\0&1&0\end{bsmallmatrix}\mathbf{K}_j(\mathbf{f}_i)_j$, where $\mathbf{K}_j$ is the camera calibration matrix of $c_j$ and $(\mathbf{f}_i)_j=[(x_i)_j/(z_i)_j, (y_i)_j/(z_i)_j,1]^\top$ is the normalized image coordinates of $(\mathbf{x}_i)_j$.
Then, $(\mathbf{f}_i)_j$ can be obtained by $(\mathbf{f}_i)_j=\mathbf{K}_j^{-1}[(u_i)_j, (v_i)_j, 1]^\top$.  
We denote the rotation and the translation between camera $j$ and $k$ as $\mathbf{R}_{jk}$ and $\mathbf{t}_{jk}$. 
The point $\mathbf{x}_i$ in the reference frame of $c_j$ and $c_k$ is related by $(\mathbf{x}_i)_j=\mathbf{R}_{jk}(\mathbf{x}_i)_k+\mathbf{t}_{jk}$.
This means that $\mathbf{R}_{jk}=\mathbf{R}_{j}\mathbf{R}_{k}^\top$ and $\mathbf{t}_{jk}=-\mathbf{R}_{j}\mathbf{R}_{k}^\top\mathbf{t}_{k}+\mathbf{t}_{j}$.

\section{Review of Two-view Rotation-Only Method}
\label{sec:review_kneip}
In this section, we review the two-view rotation-only optimization method proposed by Kneip and Lynen in \cite{kneip_rotation_ICCV}.
Consider two views with known internal calibration, $c_j$ and $c_k$, observing $m\geq5$ common points with index $i \in \{1, 2, \dots, m\}$. 
The normalized epipolar error \cite{nee} associated with each point $i$ is defined as
\begin{equation}
\label{eq:nee}
    (e_i)_{(j,k)} = \left|\hspace{1pt} \widehat{\mathbf{t}}_{jk}\cdot\left(\big(\hspace{2pt}\widehat{\mathbf{f}}_i\big)_j\times\mathbf{R}_{jk}\big(\hspace{2pt}\widehat{\mathbf{f}}_i\big)_k\right)\right|,
\end{equation}
where $\big(\hspace{2pt}\widehat{\mathbf{f}}_i\big)_j$ and $\big(\hspace{2pt}\widehat{\mathbf{f}}_i\big)_k$ are the unit bearing vectors corresponding to the $i$th point in $c_j$ and $c_k$, respectively. 
The sum of squares of all these errors is then given by
\begin{equation}
\label{eq:sos_nee}
    \sum_{i=1}^m (e_i)_{(j,k)}^2 = \widehat{\mathbf{t}}_{jk}^{\hspace{1pt}\top}\mathbf{M}_{jk}\widehat{\mathbf{t}}_{jk},
    \vspace{-1.5em}
\end{equation}
where 
\begin{flalign}
\label{eq:M_jk}
    &\mathbf{M}_{jk}{\hspace{1pt}=}\sum_{i=1}^m\hspace{-2pt}\left(\hspace{-1pt}\big(\hspace{2pt}\widehat{\mathbf{f}}_i\big)_j{\times\hspace{1pt}}\mathbf{R}_{jk}\big(\hspace{2pt}\widehat{\mathbf{f}}_i\big)_k\right)\hspace{-2pt}\left(\hspace{-1pt}\big(\hspace{2pt}\widehat{\mathbf{f}}_i\big)_j{\times\hspace{1pt}}\mathbf{R}_{jk}\big(\hspace{2pt}\widehat{\mathbf{f}}_i\big)_k\right)^\top\hspace{-5pt}. \hspace{-2em}&
\end{flalign}
In \cite{kneip_rotation_ICCV}, it was shown that the $3\times3$ matrix $\mathbf{M}_{jk}$ can also be computed as follows:
denoting the entries of $\big(\hspace{2pt}\widehat{\mathbf{f}}_i\big)_j$ as
$[(f_{xi})_j, (f_{yi})_j, (f_{zi})_j]^\top$, the following matrices are defined:
\begin{align}
    (\mathbf{F}_{xx})_{jk} &= {\textstyle\sum_{i=1}^m} (f_{xi})_j^2\big(\hspace{2pt}\widehat{\mathbf{f}}_i\big)_k\big(\hspace{2pt}\widehat{\mathbf{f}}_i\big)_k^\top, \label{eq:Fxx}\\  
    (\mathbf{F}_{xy})_{jk} &= {\textstyle\sum_{i=1}^m} (f_{xi})_j(f_{yi})_j\big(\hspace{2pt}\widehat{\mathbf{f}}_i\big)_k\big(\hspace{2pt}\widehat{\mathbf{f}}_i\big)_k^\top, \\
    (\mathbf{F}_{xz})_{jk} &= {\textstyle\sum_{i=1}^m} (f_{xi})_j(f_{zi})_j\big(\hspace{2pt}\widehat{\mathbf{f}}_i\big)_k\big(\hspace{2pt}\widehat{\mathbf{f}}_i\big)_k^\top, \\
    (\mathbf{F}_{yy})_{jk} &= {\textstyle\sum_{i=1}^m} (f_{yi})_j^2\big(\hspace{2pt}\widehat{\mathbf{f}}_i\big)_k\big(\hspace{2pt}\widehat{\mathbf{f}}_i\big)_k^\top, \\  
    (\mathbf{F}_{yz})_{jk} &= {\textstyle\sum_{i=1}^m} (f_{yi})_j(f_{zi})_j\big(\hspace{2pt}\widehat{\mathbf{f}}_i\big)_k\big(\hspace{2pt}\widehat{\mathbf{f}}_i\big)_k^\top. \label{eq:Fyz}
\end{align}
Let $\mathbf{r}_1$, $\mathbf{r}_2$, $\mathbf{r}_3$
be each row of $\mathbf{R}_{jk}$, and $m_{ab}$ be the element of $\mathbf{M}_{jk}$ at the $a$th row and $b$th column.
(Notice that we omitted the subscript $jk$ here for simplicity).
Then,
\begin{flalign}
    m_{11}&=
    \mathbf{r}_3\mathbf{F}_{yy}\mathbf{r}_3^\top
    -2\mathbf{r}_3\mathbf{F}_{yz}\mathbf{r}_2^\top
    +\mathbf{r}_2\mathbf{F}_{zz}\mathbf{r}_2^\top, &\label{eq:m11}\\
    m_{22}&=
    \mathbf{r}_1\mathbf{F}_{zz}\mathbf{r}_1^\top
    -2\mathbf{r}_1\mathbf{F}_{xz}\mathbf{r}_3^\top
    +\mathbf{r}_3\mathbf{F}_{xx}\mathbf{r}_3^\top, &\\
    m_{33}&=
    \mathbf{r}_2\mathbf{F}_{xx}\mathbf{r}_2^\top
    -2\mathbf{r}_1\mathbf{F}_{xy}\mathbf{r}_2^\top
    +\mathbf{r}_1\mathbf{F}_{yy}\mathbf{r}_1^\top,&\\
    m_{12}&=
    \mathbf{r}_1\mathbf{F}_{yz}\mathbf{r}_3^\top
    {-\hspace{2pt}}\mathbf{r}_1\mathbf{F}_{zz}\mathbf{r}_2^\top
    {-\hspace{2pt}}\mathbf{r}_3\mathbf{F}_{xy}\mathbf{r}_3^\top
    {+\hspace{2pt}}\mathbf{r}_3\mathbf{F}_{xz}\mathbf{r}_2^\top,\hspace{-1em} &\\
    m_{13}&=
    \mathbf{r}_2\mathbf{F}_{xy}\mathbf{r}_3^\top
    {-\hspace{2pt}}\mathbf{r}_2\mathbf{F}_{xz}\mathbf{r}_2^\top
    {-\hspace{2pt}}\mathbf{r}_1\mathbf{F}_{yy}\mathbf{r}_3^\top
    {+\hspace{2pt}}\mathbf{r}_1\mathbf{F}_{yz}\mathbf{r}_2^\top,\hspace{-1em} & \\
     m_{23}&=
    \mathbf{r}_1\mathbf{F}_{xz}\mathbf{r}_2^\top
    {-\hspace{2pt}}\mathbf{r}_1\mathbf{F}_{yz}\mathbf{r}_1^\top
    {-\hspace{2pt}}\mathbf{r}_3\mathbf{F}_{xx}\mathbf{r}_2^\top
    {+\hspace{2pt}}\mathbf{r}_3\mathbf{F}_{xy}\mathbf{r}_1^\top,\hspace{-1em} & \label{eq:m23}
    \end{flalign}
and $m_{21}=m_{12}$, $m_{31}=m_{13}$, $m_{32}=m_{23}$.
This is a more efficient computation of $\mathbf{M}_{jk}$ than \eqref{eq:M_jk}, as \eqref{eq:Fxx}--\eqref{eq:Fyz} can be precomputed regardless of $\mathbf{R}_{jk}$, reducing the number of operations during the rotation optimization \cite{kneip_rotation_ICCV}.

Given the set of corresponding unit bearing vectors, one can jointly optimize the relative rotation and translation by minimizing \eqref{eq:sos_nee} with respect to $\mathbf{R}_{jk}$ and $\widehat{\mathbf{t}}_{jk}$.
In \cite{kneip_rotation_ICCV}, it was shown that this problem can be transformed into a rotation-only form:
\begin{equation}
\label{eq:optimization_kneip}
    \mathbf{R}_{jk}^*=\argmin_{\mathbf{R}_{jk}} \lambda_\mathbf{M}(\mathbf{R}_{jk}),
\end{equation}
where $\lambda_\mathbf{M}(\mathbf{R}_{jk})$ is the smallest eigenvalue of $\mathbf{M}_{jk}$ (which is a function of $\mathbf{R}_{jk}$).
This eigenvalue can be obtained in closed form \cite{kneip_rotation_ICCV}:
\begin{align}
    b_1 &= -m_{11}-m_{22}-m_{33}, \label{eq:b1}\\
    b_2 &= -m_{13}^2-m_{23}^2-m_{12}^2 \nonumber\\
    &\hspace{1em}+m_{11}m_{22}+m_{11}m_{33}+m_{22}m_{33},\\
    b_3 &= m_{22}m_{13}^2+m_{11}m_{23}^2+m_{33}m_{12}^2 \nonumber\\
    &\hspace{1em}-m_{11}m_{22}m_{33} -2m_{12}m_{23}m_{13}, \\
    s &= 2b_1^3-9b_1b_2+27b_3, \\
    t &= 4(b_1^2-3b_2)^3,\\
    k &= \left(\sqrt{t}/2\right)^{1/3}\cos\left(\arccos\left(s/\sqrt{t}\right)/3\right),\\
    \lambda_\mathbf{M}(\mathbf{R}_{jk}) &= \left(-b_1-2k\right)/3.\label{eq:lambda}
\end{align}
To summarize, the rotation part of the optimal solution $(\mathbf{R}_{jk}^*, \widehat{\mathbf{t}}_{jk}^*)$ that minimizes \eqref{eq:sos_nee} is obtained by solving \eqref{eq:optimization_kneip}.

\section{Rotation-Only Bundle Adjustment}
\label{sec:method}
\subsection{Cost function}
We extend the idea of \cite{kneip_rotation_ICCV} for $n$ views. Let $\mathcal{E}$ be the set of all edges, \ie, camera pairs $(j, k)$ observing a sufficient number of points in common ($>10$ in our implementation).
Then, we formulate the optimization problem as follows:
\begin{equation}
\label{eq:our_cost1}
\{\mathbf{R}_1^*, \cdots, \mathbf{R}_n^*\}
=\argmin_{\mathbf{R}_1, \cdots, \mathbf{R}_n}C(\mathbf{R}_1, \cdots, \mathbf{R}_n)
\vspace{-0.5em}
\end{equation}
with
\vspace{-0.5em}
\begin{equation}
\label{eq:our_cost2}
     C(\mathbf{R}_1, \cdots, \mathbf{R}_n)= \sum_{(j,k)\in\mathcal{E}} \underbrace{\sqrt{\lambda_\mathbf{M}(\mathbf{R}_{jk})}}_{c_{jk}},
\end{equation}
where $\lambda_\mathbf{M}(\mathbf{R}_{jk})$ is the same cost function used in \eqref{eq:optimization_kneip} for the two-view case and $c_{jk}=\sqrt{\lambda_\mathbf{M}(\mathbf{R}_{jk})}$ is our edge cost.
We empirically found that this square rooting improves the convergence rate (see Table \ref{tab:ablation1}), which we presume is due to the downweighted influence of outliers.
Alg. \ref{al:edge_cost} summarizes the steps for computing the edge cost $c_{jk}$.

\begin{algorithm}[t]
\setstretch{1}
\caption{Computation of edge cost $c_{jk}$ }
\label{al:edge_cost}
\DontPrintSemicolon
\KwInput{Relative rotation $\mathbf{R}_{jk}$, Precomputed matrices\\
\hspace{3em}$(\mathbf{F}_{xx})_{jk}, (\mathbf{F}_{xy})_{jk}, (\mathbf{F}_{xz})_{jk}, (\mathbf{F}_{yy})_{jk}, (\mathbf{F}_{yz})_{jk}.\hspace{-10em}$}
\KwOutput{Edge cost $c_{jk}$.}

$\mathbf{r}_i\gets(\mathbf{R}_{jk})_{i\text{th row}}$ for $i=1,2,3;\hspace{-5em}$

compute $m_{11}, \cdots, m_{23}$ using \eqref{eq:m11}--\eqref{eq:m23};\hspace{-5em}\label{line:M}

compute $\lambda_\mathbf{M}(\mathbf{R}_{jk})$ using \eqref{eq:b1}--\eqref{eq:lambda};

$c_{jk}\gets\sqrt{\lambda_\mathbf{M}(\mathbf{R}_{jk})};\label{line:square_rooting}$

\Return $c_{jk}$;
\end{algorithm}
\begin{algorithm}[t]
\setstretch{1}
\caption{Rotation-Only Bundle Adjustment}
\label{al:proposed}
\DontPrintSemicolon
\KwInput{Initial rotations $\mathbf{R}_1, \cdots, \mathbf{R}_n$, \ \  $\text{Edges } \mathcal{E},\hspace{-10em}$  \\ 
\hspace{3em}Matched unit bearing vectors $\big(\big(\hspace{2pt}\widehat{\mathbf{f}}_i\big)_j, \big(\hspace{2pt}\widehat{\mathbf{f}}_i\big)_k\big)$ \hspace{-5em}\\
\hspace{3em}for all $i\in1,\cdots,m_{(j,k)}$ for all $(j,k)\in\mathcal{E}$,\\
\hspace{3em}Number of iterations $n_\text{it}$.\hspace{-10em}}
\KwOutput{Final rotations $\mathbf{R}_1, \cdots, \mathbf{R}_n$.}

\tcc{Initialization:}

compute $(\mathbf{F}_{xx})_{jk}, \cdots, (\mathbf{F}_{yz})_{jk}\hspace{-5em}$ \hspace{5em}
for all $(j,k)\in\mathcal{E}$ using \eqref{eq:Fxx}--\eqref{eq:Fyz}; \label{line:init1}

compute $C$ using \eqref{eq:our_cost2} and Alg. \ref{al:edge_cost};

obtain $\mathbf{s}_0$ by stacking $\mathrm{Log}(\mathbf{R}_1), \cdots, \mathrm{Log}(\mathbf{R}_n)$ in one column;

$\beta_1{\hspace{2pt}\gets\hspace{2pt}}0.9$; \ $\beta_2{\hspace{2pt}\gets\hspace{2pt}}0.999$; \ $\alpha{\hspace{2pt}\gets\hspace{2pt}}0.01$; \ $\bm{\epsilon}{\hspace{2pt}\gets\hspace{2pt}}(10^{-8})\mathbf{1}_{3n\times1};\hspace{-5em}$

$\mathbf{m}_0\gets\mathbf{0}_{3n\times1}$; \ \  $\mathbf{v}_0\gets\mathbf{0}_{3n\times1}$; \ \ $t\gets0;\label{line:init2}$

\tcc{Optimization:}

\While{$t < n_\text{it}$ \label{line:opti1}}
{

$t\gets t+1$;

compute $C$ and $\mathbf{g}_t$ using Alg. \ref{al:cost_and_gradient};\label{line:gradient}

Perform \eqref{eq:adam_m}--\eqref{eq:adam_last};

\If{$C$ increased in five successive iterations}{$\alpha\gets0.001$;\label{line:opti2}}

}
\Return $\mathbf{R}_1, \cdots, \mathbf{R}_n$;
\end{algorithm}

\subsection{Optimization}
\label{subsec:optimization}
To solve \eqref{eq:our_cost1} iteratively, we use Adam \cite{adam}, a first-order gradient-based optimization algorithm for stochastic objective functions. 
Adam has been widely used in deep learning, and we found that it also works well for our geometric optimization problem.
Given the initial estimates of $\mathbf{R}_1, \cdots, \mathbf{R}_n$, let $\mathbf{s}_0$ be the initial state vector formed by stacking $\mathrm{Log}(\mathbf{R}_1),\cdots, \mathrm{Log}(\mathbf{R}_n)$ in one column.
Let $\mathbf{m}_0=\mathbf{0}_{3n\times1}$, $\mathbf{v}_0=\mathbf{0}_{3n\times1}$, $t=0$ and $\bm{\epsilon}=(10^{-8})\mathbf{1}_{3n\times1}$.
Then, using Adam, we repeat the following steps at each iteration $t$ of our optimization:
\begin{flalign}
    t&\gets t+1,&\\
    \mathbf{g}_t&\gets \nabla_\mathbf{s}C(\mathbf{R}_1, \cdots, \mathbf{R}_n),&\label{eq:adam_gradient}\\
    \mathbf{m}_t&\gets \beta_1\mathbf{m}_{t-1}+(1-\beta_1)\mathbf{g}_t,&\label{eq:adam_m}\\
    \mathbf{v}_t&\gets\beta_2\mathbf{v}_{t-1}+(1-\beta_2)(\mathbf{g}_t\circ\mathbf{g}_t),&\\
    \mathbf{m}'_t&\gets\mathbf{m}_t/(1-\beta_1^t),&\\
    \mathbf{v}'_t&\gets\mathbf{v}_t/(1-\beta_2^t),&\\
    \mathbf{s}_t&\gets\mathbf{s}_{t-1}-\alpha\mathbf{m}'_t\oslash(\mathbf{v}'_t{}^{\circ1/2}+\bm{\epsilon}),&\\
    \mathbf{u}_i&\gets[(\mathbf{s}_t)_{3i-2},(\mathbf{s}_t)_{3i-1},(\mathbf{s}_t)_{3i}]^\top \text{ for } i=1,\cdots,n,\hspace{-3em}&\\
    \mathbf{R}_i&\gets\mathrm{Exp}(\mathbf{u}_i) \text{ for } i = 1,\cdots, n. \label{eq:adam_last}&
\end{flalign}
We detail the computation of the gradient (\ie, \eqref{eq:adam_gradient}) in the next section.
For the hyper-parameters $\beta_1$ and $\beta_2$, we use the default values given in \cite{adam}: $\beta_1=0.9$ and $\beta_2=0.999$.
For the step size $\alpha$, we use $\alpha=0.01$ at the beginning and switch to $\alpha=0.001$ permanently once the cost increases in five successive iterations.
We empirically found that this switching sometimes helps the convergence (see Table \ref{tab:ablation1}).
Alg. \ref{al:proposed} summarizes our method.

\subsection{Gradient computation}
We compute the gradient $\mathbf{g}_t$ in \eqref{eq:adam_gradient} numerically\footnote{
It is possible to compute it analytically. 
However, the closed-form expressions involve more operations than the numerical method (see the supplementary material of \cite{kneip_rotation_ICCV}). 
We empirically found that this takes approximately 1.8 times longer, while the numerical difference is negligible.
}.
This can be done efficiently by slightly perturbing each rotation parameter in $\mathbf{s}_t$ and summing the resulting changes of all the edge costs $c_{jk}$ \eqref{eq:our_cost2} as we traverse the edge set $\mathcal{E}$.
Since we need to run Alg. \ref{al:edge_cost} seven times for each edge (\ie, 1 from the unperturbed state, $3\times2$ from perturbing $\mathbf{R}_j$ and $\mathbf{R}_k$), if there are $n_\mathcal{E}$ edges, this method will require $7n_\mathcal{E}$ computations of edge costs.
To reduce the computation time, we make the following approximation:
\begin{flalign}
    &c_{jk}\left(\mathbf{R}_j(\mathbf{R}_k)_{x+\Delta x}^\top\right) -c_{jk}(\mathbf{R}_j\mathbf{R}_k^\top)&\nonumber\\
    &\hspace{5em}\approx c_{jk}(\mathbf{R}_j\mathbf{R}_k^\top) - c_{jk}\left((\mathbf{R}_j)_{x+\Delta x}\mathbf{R}_k^\top\right),\hspace{-5em}&\label{eq:gradient_assumption} 
\end{flalign}
where $(\mathbf{R}_j)_{x+\Delta x}$ and $(\mathbf{R}_k)_{x+\Delta x}$ respectively denote $\mathbf{R}_j$ and $\mathbf{R}_k$ after being perturbed (by the same magnitude) in the $x$ component of the rotation vector. 
That is, we assume that $\Delta c_{jk}$ due to $(\mathbf{R}_k)_{x+\Delta x}$ is approximately equal to the negative of $\Delta c_{jk}$ due to $(\mathbf{R}_j)_{x+\Delta x}$.
We make analogous approximations for the perturbations in the $y$ and $z$ component of the rotation vector.
By approximating the gradient of $\mathbf{R}_k$ using that of $\mathbf{R}_j$, we reduce the number of edge cost computations from $7n_\mathcal{E}$ to $4n_\mathcal{E}$.
Empirically, we found that this improves the efficiency significantly at a relatively small loss of accuracy (see Table \ref{tab:ablation2}).
Alg. \ref{al:cost_and_gradient} summarizes the procedure for computing the gradient and the total cost.

\begin{algorithm}[t]
\setstretch{1}
\caption{Cost and gradient computation}
\label{al:cost_and_gradient}
\DontPrintSemicolon
\KwInput{Current rotations $\mathbf{R}_1, \cdots, \mathbf{R}_n$, \ \  $\text{Edges } \mathcal{E},\hspace{-10em}$  \\ 
\hspace{3em}$(\mathbf{F}_{xx})_{jk}, (\mathbf{F}_{xy})_{jk}, (\mathbf{F}_{xz})_{jk}, (\mathbf{F}_{yy})_{jk}, (\mathbf{F}_{yz})_{jk}\hspace{-10em}$\\
\hspace{3em}for all $(j,k)\in\mathcal{E}$}
\KwOutput{Cost $C$, Gradient $\mathbf{g}$.}

$C\gets0;$ \ \ $\mathbf{g}\gets\mathbf{0}_{3n\times1};$ \ \ $\delta\gets10^{-4};$

$\bm{\delta}_x\gets[\delta,0,0]^\top; \ \  \bm{\delta}_y\gets[0,\delta,0]^\top;\ \  \bm{\delta}_z\gets[0,0,\delta]^\top;\hspace{-5em}$ 

$\mathbf{u}_i\gets\mathrm{Log}(\mathbf{R}_i)$ for all $i\in1,\cdots,n;$

\tcc{Perturb the rotations:}

$(\mathbf{R}_i)_{x+\Delta x}\gets\mathrm{Exp}\left(\mathbf{u}_i+\bm{\delta}_x\right)$ for all $i\in1,\cdots,n;$

$(\mathbf{R}_i)_{y+\Delta y}\gets\mathrm{Exp}\left(\mathbf{u}_i+\bm{\delta}_y\right)$ for all $i\in1,\cdots,n;$

$(\mathbf{R}_i)_{z+\Delta z}\gets\mathrm{Exp}\left(\mathbf{u}_i+\bm{\delta}_z\right)$ for all $i\in1,\cdots,n;$

\tcc{Sum the resulting changes of each $c_{jk}:\hspace{-5em}$}

\For{$(j,k)\in\mathcal{E}$}
{
$\mathbf{R}_{jk}\gets\mathbf{R}_j\mathbf{R}_k^\top;$

$(\mathbf{R}_{jk})_{x+\Delta x}\gets(\mathbf{R}_j)_{x+\Delta x}\mathbf{R}_k^\top;$

$(\mathbf{R}_{jk})_{y+\Delta y}\gets(\mathbf{R}_j)_{y+\Delta y}\mathbf{R}_k^\top;$

$(\mathbf{R}_{jk})_{z+\Delta z}\gets(\mathbf{R}_j)_{z+\Delta z}\mathbf{R}_k^\top;$

obtain $c_{jk}$ using Alg. \ref{al:edge_cost} with $\mathbf{R}_{jk}$.

obtain $(c_{jk})_{x+\Delta x}$ using Alg. \ref{al:edge_cost} with $(\mathbf{R}_{jk})_{x+\Delta x}$.

obtain $(c_{jk})_{y+\Delta y}$ using Alg. \ref{al:edge_cost} with $(\mathbf{R}_{jk})_{y+\Delta y}$.

obtain $(c_{jk})_{z+\Delta z}$ using Alg. \ref{al:edge_cost} with $(\mathbf{R}_{jk})_{z+\Delta z}$.

$\Delta(c_{jk})_x\gets(c_{jk})_{x+\Delta x}-c_{jk};$

$\Delta(c_{jk})_y\gets(c_{jk})_{y+\Delta y}-c_{jk};$

$\Delta(c_{jk})_z\gets(c_{jk})_{z+\Delta z}-c_{jk};$

$\mathbf{g}_{3j-2}\gets \mathbf{g}_{3j-2}+\Delta(c_{jk})_x;$

$\mathbf{g}_{3j-1}\gets \mathbf{g}_{3j-1}+\Delta(c_{jk})_y;$

$\mathbf{g}_{3j}\gets \mathbf{g}_{3j}+\Delta(c_{jk})_z;$

$\mathbf{g}_{3k-2}\gets \mathbf{g}_{3k-2}-\Delta(c_{jk})_x;\label{line:g_k1}$

$\mathbf{g}_{3k-1}\gets \mathbf{g}_{3k-1}-\Delta(c_{jk})_y;$

$\mathbf{g}_{3k}\gets \mathbf{g}_{3k}-\Delta(c_{jk})_z;\label{line:g_k2}$

$C\gets C+ c_{jk};$
}

$\mathbf{g}\gets\mathbf{g}/\delta;$

\Return $C$ and $\mathbf{g}$;
\end{algorithm}

\section{Results}
\label{sec:results}
\vspace{-0.1em}
\subsection{Evaluation method}
\label{subsec:eval_method}
We compare our method (henceforth ROBA) against the state-of-the-art rotation averaging method by Chatterjee and Govindu \cite{chatterjee2018robust} (henceforth RA).
Both methods are implemented in MATLAB and run on a laptop CPU (Intel i7-4710MQ, 2.8GHz).
For RA, we use the code publicly shared by the authors of \cite{chatterjee2018robust}\footnote{\url{http://www.ee.iisc.ac.in/labs/cvl/research/rotaveraging/}} with the $L_{\frac{1}{2}}$ loss function, as recommended in \cite{chatterjee2018robust}.
We use the output of RA as input to ROBA, so that we can compare RA versus RA $+$ ROBA.

In Alg. \ref{al:proposed}, the bottleneck is the gradient computation (line \ref{line:gradient}), where the predominant part is the computation of the edge costs using Alg. \ref{al:edge_cost}.
To speed up this part, we implement Alg. \ref{al:edge_cost} in a C++ MEX function.
We set the number of iterations $(n_\text{it})$ to 100 in Alg. \ref{al:proposed}.
Note that, in practice, it would be sensible to adopt some stopping criteria to detect the convergence (\eg, based on the relative change of the total cost or the angular change in the rotations).
In our experiment, however, we aim to investigate the convergence behavior of ROBA, so we are agnostic about this heuristics.

Finally, we draw attention to the error metrics for evaluating the rotation estimates $(\mathbf{R}_1, \cdots, \mathbf{R}_n)$.
Since they do not share the same reference frame as the ground-truth rotations $(\mathbf{R}_1^\text{gt}, \cdots, \mathbf{R}_n^\text{gt})$, we must first align them with the ground truth to evaluate the accuracy.
Commonly, this is done by rotating them with one of the following rotations:
\begin{align}
    \mathbf{R}_{L1} &= \argmin_{\mathbf{R}_{L1}}\sum_{j=1}^n d\left(\mathbf{R}_{L1}, \mathbf{R}_j^\top\mathbf{R}_j^\text{gt}\right),\label{eq:R_L1}\\
    \mathbf{R}_{L2} &=
    \argmin_{\mathbf{R}_{L2}}\sum_{j=1}^n d\left(\mathbf{R}_{L2},\mathbf{R}_j^\top\mathbf{R}^\text{gt}_j\right)^2,\label{eq:R_L2}\vspace{-1em}
\end{align}
where $d(\cdot, \cdot)$ denotes the geodesic distance between the two rotations, \ie, $d(\mathbf{R}_1, \mathbf{R}_2)=\arccos((\mathrm{tr}(\mathbf{R}_1\mathbf{R}_2^\top)-1)/2)$.
Note that \eqref{eq:R_L1} and \eqref{eq:R_L2} are single rotation averaging problems, and they can be solved using iterative algorithms \cite{hartley2011L1, hartley_rotation_averaging}.
Afterwards, we rotate the estimates as follows\footnote{We must use right multiplication here in order to make sure that $(\mathbf{R}_{12})_\text{after}=(\mathbf{R}_{1}\mathbf{R}_{L1})(\mathbf{R}_{2}\mathbf{R}_{L1})^\top=\mathbf{R}_1\mathbf{R}_2^\top=(\mathbf{R}_{12})_\text{before}$.}:
\vspace{-0.5em}
\begin{equation}
\label{eq:alignment}
    \mathbf{R}_j\gets\mathbf{R}_j\mathbf{R}_{L1}, \quad\text{ or}\quad 
    \mathbf{R}_j\gets\mathbf{R}_j\mathbf{R}_{L2}.
\end{equation}
Since $\mathbf{R}_{L1}$ minimizes the sum of absolute distances and $\mathbf{R}_{L2}$ minimizes the sum of squares, we call the first method $L_1$ \textit{alignment} and the second $L_2$ \textit{alignment}.
In our evaluation, we report the mean and median angular errors using these two alignment methods.

\begin{figure*}[t]
    \centering
    \includegraphics[width=0.9\textwidth]{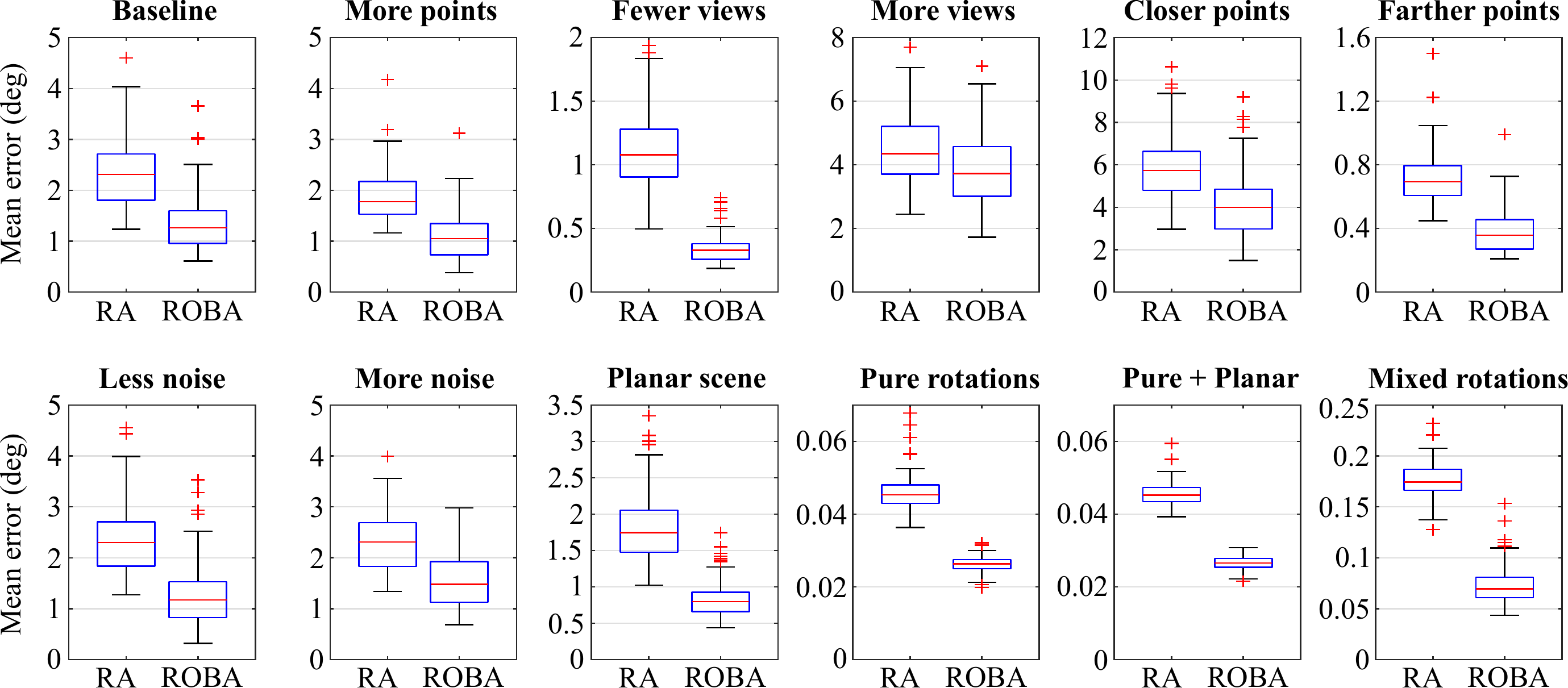}
    \vspace{0.5em}
    \caption{
    Results on the synthetic data (see Table \ref{tab:simulation_settings} for the settings).
    We compare RA \cite{chatterjee2018robust} and ROBA (initialized by RA) in terms of the mean angular error after the $L_1$ alignment (see the supplementary material for the other error metrics).
    It shows that ROBA improves the results of RA in all scenarios considered.
    In particular, the relative error reduction is large for fewer views, farther points and pure rotations, all of which lead to a denser view-graph (see Table \ref{tab:synthetic_result}).
    }
    \label{fig:synthetic_result}
    \vspace{-0.5em}
\end{figure*}

\begin{figure}[t]
    \centering
    \includegraphics[width=0.47\textwidth]{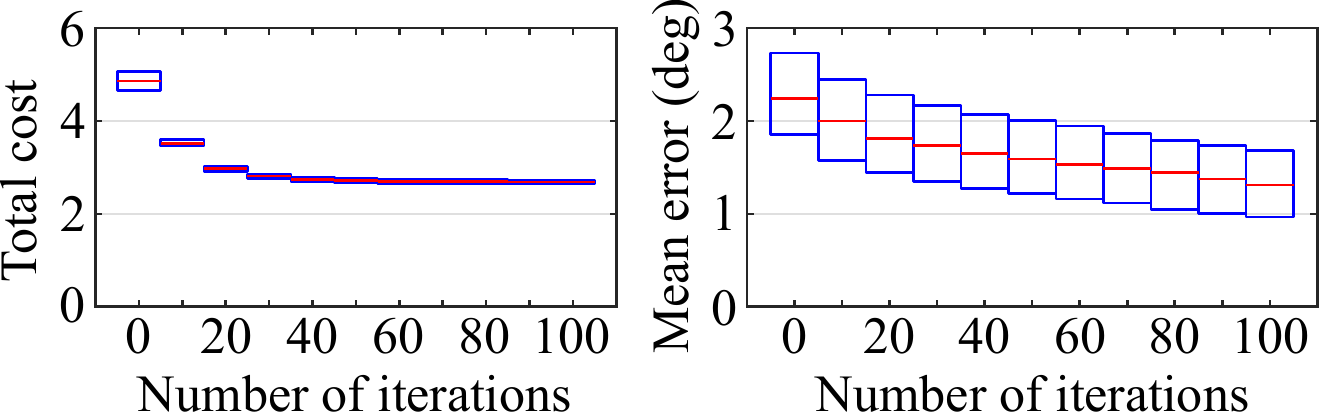}
    \vspace{0.5em}
    \caption{
    Evolution of the cost function (left) and the mean angular errors after the $L_1$ alignment (right) in the baseline setting.
    Here, we only show the interquartile range.
    Notice that while our cost \eqref{eq:our_cost2} seems to plateau after 30--40 iterations, the actual rotation errors continue to decrease. 
    This is discussed in Section \ref{subsec:convergence}.
    }
    \label{fig:synthetic_evolution}
\end{figure}

\subsection{Synthetic data}
To study how different factors affect our method, we run Monte Carlo simulations in controlled settings:
we uniformly distribute $n$ cameras on a circle on the xy-plane such that the neighbors are $1$ unit apart.
After aligning their optical axes with the z-axis, we perturb the rotations by random angles $\theta\sim\mathcal{U}(0, 20^\circ)$. 
We set the image size to $640\times480$ pixels and the focal length to $525$ pixels, the same as those in \cite{tum_rgbd}.
We create 3D points at random distances $d\sim\mathcal{U}(d_\text{min}, d_\text{max})$ from the xy-plane, ensuring that every neighboring view observes at least $n_\text{cov}$ points in common.
We perturb the image coordinates of the points by $\mathcal{N}(0, \sigma^2)$.
For every pair of views observing $n_\text{cov}$ or more points in common, we estimate the relative pose and the inlying points.
If there are at least 10 inliers, we add the pair as an edge in $\mathcal{E}$.
Table \ref{tab:simulation_settings} specifies the configuration parameters we set for our simulations.
For each parameter setting, we generate 100 different datasets, each with randomly sampled camera rotations, 3D points and 2D measurements.

To obtain the relative rotation estimates, we use the following method:
first, we obtain 100 pose samples around the ground-truth relative pose.
We do this by perturbing the rotation and the translation by two arbitrary angles ($<20$ deg).
Then, we evaluate the $L_1$-optimal angular reprojection errors \cite{lee2019closed}\footnote{This can be computed using \eqref{eq:nee} and their relation derived in \cite{nee}. We do not consider cheirality \cite{hartley_book}, because otherwise we would end up discarding many inlying low-parallax points that appear in pure rotations.} for each pose and choose the one that yields the most inliers.
This method is similar to the standard method of using a minimal solver (\eg, five-point algorithm \cite{nister_five_point_PAMI}) in RANSAC \cite{ransac}, except that our samples are simulated, not estimated.
We use this method because our focus is on the optimization of the rotations and we are agnostic about the relative pose estimation method.

Fig. \ref{fig:synthetic_result} and Table \ref{tab:synthetic_result} present the results in each setting. 
Notice that ROBA improves the results of RA in all scenarios considered.
In Fig. \ref{fig:synthetic_evolution}, we show the evolution of our cost function \eqref{eq:our_cost2} and the rotation error in the baseline setting. 
\vspace{-0.3em}

\begin{table}[t]
\small
\vspace{-1.5em}
\begin{center}
\setlength{\tabcolsep}{3pt}
    \begin{tabular}{c|l}
    \thickhline
    \rule{0pt}{1em}Baseline & $n = 100$, $n_\text{cov} = 50$, $\sigma = 1$px, $d_\text{min} = 2$, $d_\text{max} = 5$,\\
    setting  & Views are uniformly spaced by $1$ unit on a circle.\\
    \thickhline
    \multirow{9}{*}{Variations} & \rule{0pt}{1em}\textbf{More points} ($n_\text{cov} = 100$), \textbf{Fewer views} ($n = 30$), \\
    & \textbf{More views} ($n = 300$), \textbf{Closer points} ($d_\text{max}=3$), \\
    & \textbf{Farther points} ($d_\text{max}{=}10$), \textbf{Less noise} ($\sigma{\hspace{2pt}=\hspace{2pt}}0.5$px),\\
    & \textbf{More noise} ($\sigma=2$px), \textbf{Planar scene} ($d_\text{min} = 5$).\\
    & \textbf{Pure rotations} (all views are placed at the origin),\\
    & \textbf{Pure rotations $\bm{+}$ Planar scene} ($d_\text{min} = 5$),\\
    & \textbf{Mixed rotations} (20 groups of views are uniform- \\
    & ly spaced by 1 unit on a circle. A group consists of \\
    & five views at the same location).\\
    \thickhline
    \end{tabular}
    \vspace{-1.2em}
    \end{center}
    \caption{Simulation settings.}
    \label{tab:simulation_settings}
    \vspace{-0.3em}
\end{table}

\begin{table}[t]
\small
\vspace{-1em}
\begin{center}
\setlength{\tabcolsep}{3pt}
\begin{tabular}{l|rrrccr}
    \thickhline
    \multirow{2}{*}{\rule{0pt}{1em}Settings} & \multirow{2}{*}{$\%_\mathcal{E}$} & \multirow{2}{*}{$\overline{e}_\mathcal{E}$} 
     & \multirow{2}{*}{$\widetilde{e}_\mathcal{E}$} & \multirow{2}{*}{RA \cite{chatterjee2018robust}} & RA \cite{chatterjee2018robust}& \multirow{2}{*}{$\%_\text{better}$}\\
     &&&&& + ROBA &\\
     \thickhline
     \rule{0pt}{1em}Baseline & 6.0\% & 2.21 & 1.39 & 2.31 & \textbf{1.26}& 100\%\\
     More points & 7.2\% & 2.51 & 1.52 & 1.78 & \textbf{1.05}& 100\%\\
     Fewer views & 21\% & 2.23 & 1.38 & 1.08 & \textbf{0.33}& 100\%\\
     More views & 2.0\% & 2.24 & 1.40 & 4.35 & \textbf{3.73}& 100\%\\
     Closer points &4.0\% & 2.57 & 1.65 & 5.74 & \textbf{3.99}& 100\%\\
     Farther points & 10\% & 1.97 & 1.22 & 0.69 & \textbf{0.36}& 100\%\\
     Less noise & 6.0\% & 2.25 & 1.40 &2.30 & \textbf{1.17}& 100\%\\
     More noise & 5.9\% & 2.23 & 1.38 &2.31 & \textbf{1.48}& 100\%\\
     Planar scene & 7.4\% & 2.61 & 1.56 & 1.74 & \textbf{0.80}& 100\%\\
     Pure rotations & 100\% & 0.89 & 0.74 & 0.045 & \textbf{0.026}& 100\%\\
     Pure $+$ Planar & 100\% & 0.89 & 0.74 &0.045&\textbf{0.027}& 100\%\\
     Mixed rotations & 37\% & 3.13 & 1.66 & 0.17& \textbf{0.069}& 100\%\\
     \thickhline
     \multicolumn{7}{c}{\rule{0pt}{1em}$\%_\mathcal{E}$, $\%_\text{better}$: proportion of existing edges and improved results,}\\
     \multicolumn{7}{c}{$\overline{e}_\mathcal{E}$, $\widetilde{e}_\mathcal{E}$: mean and median angular errors (in deg) of}\\
     \multicolumn{7}{c}{the relative rotations from all edges.}\\
     \thickhline
\end{tabular}
\vspace{-1.2em}
\end{center}
\caption{Median results of the 100 simulations in each setting.
The 5th and 6th columns give the median errors shown in Fig. \ref{fig:synthetic_result}.
Overall, the denser the view-graph, the more accurate both methods are.
ROBA improves upon RA in 1200 out of 1200 simulations.
}
\label{tab:synthetic_result}
\vspace{-5em}
\end{table}

\begin{table*}[t]
\small
\begin{center}
    \setlength{\tabcolsep}{0.5em}
    \begin{tabular}{lr|rrrr|rrrc|rrrr}
        \thickhline
        \multicolumn{2}{c|}{\rule{0pt}{1em}Datasets} & \multicolumn{4}{c|}{RA \cite{chatterjee2018robust}} & \multicolumn{4}{c|}{RA \cite{chatterjee2018robust} + ROBA (100 iter)} & \multicolumn{4}{c}{Computation time (s)}\\
        Name & $(n, n_\mathcal{E}, \%_\mathcal{E})$ & mn1 & md1 & mn2 & md2 & mn1 & md1 & mn2 & md2 & RA \cite{chatterjee2018robust} & Init & Opti & Total \\
        \thickhline
        \rule{0pt}{1.1em}ALM & (577, 96653, 58\%) & 4.08 & 1.11 & 4.74 & 2.17 & \textbf{2.35} & \textbf{0.42} & \textbf{3.20} & \textbf{1.47} & 16 & 35 & 216 & 267 \\
        ELS & (227, 18709, 73\%) & 2.10 & 0.50 & 2.37 & 0.93 & \textbf{1.06} & \textbf{0.10} & \textbf{1.48} & \textbf{0.57} & 1 & 5 & 42 & 48 \\
        GDM & (677, 33662, 15\%) & 6.05 & 2.78 & 6.14 & 3.12 & \textbf{2.43} & \textbf{1.13} & \textbf{2.44} & \textbf{1.15} & 3 & 8 & 76 & 87 \\
        MDR & (341, 23228, 40\%) & 6.20 & 1.27 & 7.23 & 2.88 & \textbf{4.42} & \textbf{0.61} & \textbf{5.71} & \textbf{2.50} & 2 & 7 & 52 & 61 \\
        MND & (480, 51172, 45\%) & 1.46 & 0.51 & 1.59 & 0.72 & \textbf{0.82} & \textbf{0.26} & \textbf{1.01} & \textbf{0.50} & 4 & 24 & 114 & 142 \\
        NTD & (553, 96672, 63\%) & 2.08 & 0.64 & 2.31 & 0.88 & \textbf{1.27} & \textbf{0.30} & \textbf{1.47} & \textbf{0.57} & 14 & 60 & 217 & 291 \\
        NYC & (332, 18787, 34\%) & 2.87 & 1.32 & 2.99 & 1.40 & \textbf{1.03} & \textbf{0.20} & \textbf{1.20} & \textbf{0.45} & 1 & 6 & 42  & 49 \\
        PDP & (338, 24121, 42\%) & 3.86 & 0.91 & 4.67 & 2.48 & \textbf{2.18} & \textbf{0.33} & \textbf{2.92} & \textbf{1.64} & 1 & 6 & 54  & 61 \\
        PIC & (2151, 275895, 12\%) & 4.14 & 2.28 & 4.18 & 2.38 & \textbf{1.58} & \textbf{0.29} & \textbf{1.75} & \textbf{0.49} & 220 & 62 & 617  & 899 \\
        ROF & (1083, 68379, 12\%) & 2.94 & 1.52 & 2.99 & 1.56 & \textbf{2.18} & \textbf{0.27} & \textbf{2.28} & \textbf{0.61} & 6 & 26 & 154  & 186 \\
        TOL & (472, 23379, 21\%) & 3.83 & 2.33 & 3.85 & 2.38 & \textbf{1.15} & \textbf{0.16} & \textbf{1.20} & \textbf{0.24} & 1 & 10 & 53  & 64 \\
        TFG & (5057, 663755, 5\%) & 3.40 & 2.34 & 3.42 & 2.28 & \textbf{2.76} & \textbf{2.09} & \textbf{2.78} & \textbf{1.79} & 553 & 153 & 1488  & 2194 \\
        USQ & (787, 23639, 8\%) & 5.59 & 4.03 & 5.60 & 4.06 & \textbf{3.26} & \textbf{0.90} & \textbf{3.46} & \textbf{1.26} & 1 & 7 & 54  & 62 \\
        VNC & (836, 98999, 28\%) & 6.12 & 1.33 & 7.96 & 4.06 & \textbf{4.96} & \textbf{0.25} & \textbf{7.31} & \textbf{3.47} & 15 & 53 & 221  & 289 \\
        YKM & (437, 27039, 28\%) & 3.67 & 1.60 & 3.72 & 1.61 & \textbf{1.66} & \textbf{0.19} & \textbf{1.74} & \textbf{0.28} & 1 & 12 & 61  & 74 \\
        \thickhline
        \multicolumn{14}{c}{\rule{0pt}{1em}$n$: \# connected views with known ground truth, $n_\mathcal{E}$: \# edges with at least 10 covisible 3D points, $\%_\mathcal{E}=n_\mathcal{E}/n\mathrm{C}2$ in \%,}\\
        \multicolumn{14}{c}{mn/md/1/2: mean/median angular error (in deg) after the $L_1$/$L_2$ alignment,}\\
        \multicolumn{14}{c}{Init: Initialization (line \ref{line:init1}--\ref{line:init2} of Alg. \ref{al:proposed}), Opti: Optimization (line \ref{line:opti1}--\ref{line:opti2} of Alg. \ref{al:proposed}). }\\
        \thickhline
    \end{tabular}
    \vspace{-0.5em}
    \end{center}
    \caption{
    Results on the real data \cite{wilson20141dsfm}.
    For all datasets, ROBA improves the results of RA \cite{chatterjee2018robust}.
    This is shown across all error criteria, and often, the relative error reduction is significant.
    See the supplementary material for the evolution of the total cost \eqref{eq:our_cost2} and the errors.
    }
    \label{tab:real_result}
\end{table*}

\newpage
\vfill\null

\subsection{Real data}
\label{subsec:real_results}
\vspace{-0.4em}
We also perform an evaluation on the real-world datasets publicly shared by the authors of \cite{wilson20141dsfm}\footnote{\url{http://www.cs.cornell.edu/projects/1dsfm/}}, which include:
\begin{itemize}
    \item internal camera calibration and radial distortion data,
    \item SIFT \cite{sift} feature tracks and their image coordinates, 
    \item estimated relative rotations,
    \item reconstruction made with \texttt{Bundler} \cite{bundler}, consisting of the camera poses and a sparse set of 3D points.\footnote{The pose estimates are not available for some of the cameras. Also, some of the SIFT features are not associated with the available 3D points.}
\end{itemize}
As in \cite{robust_single_rotation_averaging}, we use the provided reconstruction model as the ground truth in our experiment.
We undistort all image measurements when we process the input.
Additionally, some preprocessing is required for these datasets because (1) they do not provide the SIFT IDs of the reconstructed 3D points, and (2) some edges (\ie, camera pairs for which the estimated relative rotations are given) lack covisible 3D points because some of the tracks were disregarded during the reconstruction process (\eg, outliers and low-parallax points).

\begin{figure}[t]
\vspace{-0.5em}
    \centering
    \includegraphics[width=0.35\textwidth]{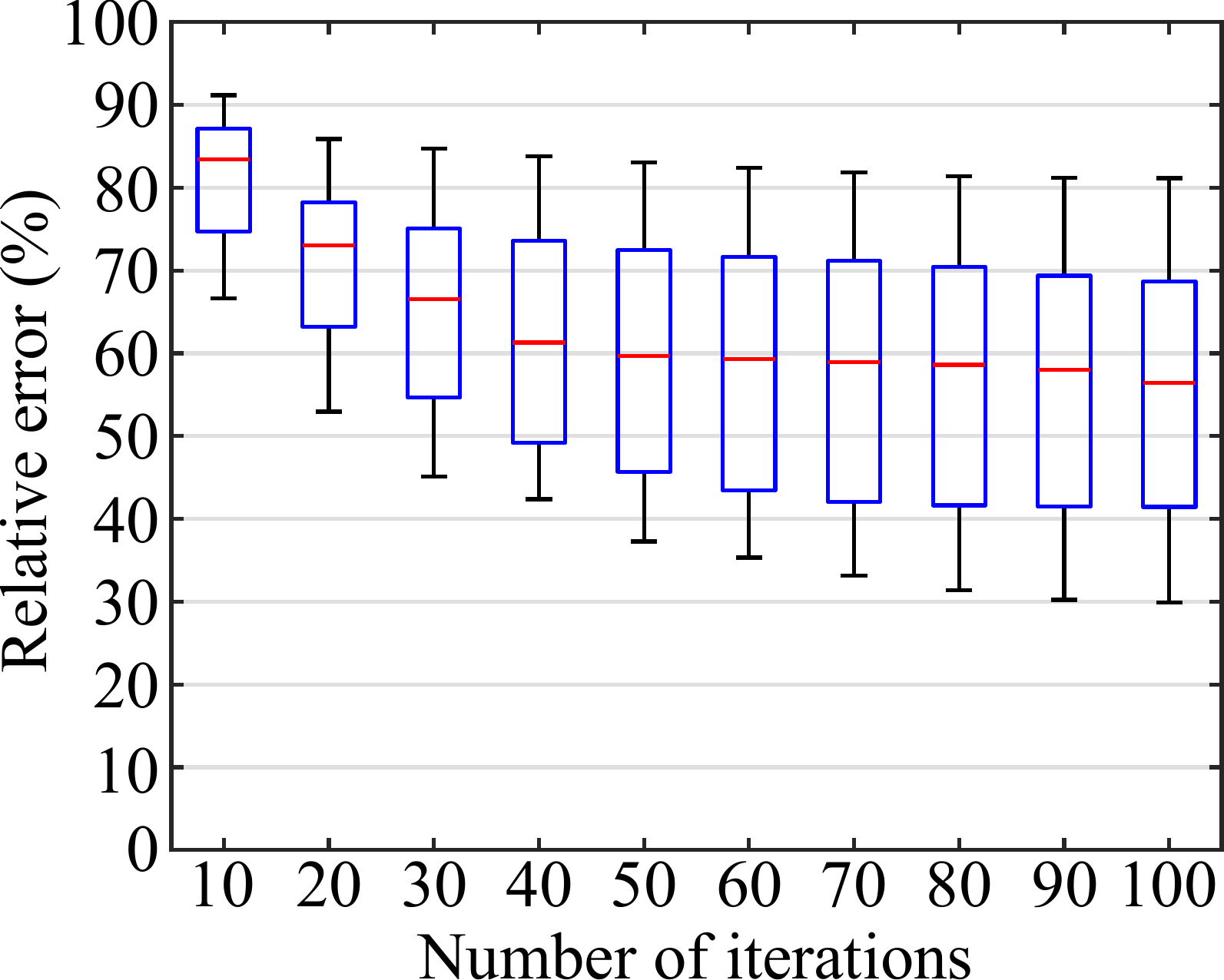}
    \caption{
    Relative errors with respect to the initial values for the real-world datasets \cite{wilson20141dsfm}.
    Here, we use the mean angular errors after the $L_1$ alignment (see the supplementary material for the other metrics).
    For more than half of the datasets, the error decreases to less than $60\%$ of the initial value after 50 iterations.  
    After 100 iterations, we observe error reductions up to 70\%.
    }
    \label{fig:evolution}
\end{figure}

We address the first issue by projecting each 3D point in its associated images and finding the track that yields the smallest mean reprojection error. 
Since most reconstructed points have small reprojection errors ($<$1--2 pixels), this can associate the points quite accurately, which we verified qualitatively.
We address the second issue by removing all edges with less than 10 covisible 3D points.
As a result, some cameras get disconnected from the main view-graph.
In our experiment, we disregard all such cameras, as well as those without the available ground truth. 

Table \ref{tab:real_result} presents the results.
It shows that ROBA offers a consistent and significant gain in accuracy.
In Fig. \ref{fig:evolution}, we plot the evolution of the relative errors aggregated from all datasets.
Table \ref{tab:ablation1} and \ref{tab:ablation2} present the ablation study results.

\begin{table}[t]
\small
\begin{center}
    \begin{tabular}{l|cc|cc|cc}
        \thickhline
        \rule{0pt}{1em} & \multicolumn{2}{c|}{\multirow{2}{*}{Baseline}} & \multicolumn{2}{c|}{Without} & \multicolumn{2}{c}{Without}\\
        & \multicolumn{2}{c|}{} & \multicolumn{2}{c|}{$\sqrt{\phantom{1}}$ in \eqref{eq:our_cost2}} & \multicolumn{2}{c}{switching $\alpha$}\\
        Datasets&  mn1 & mn2 & mn1 & mn2 & mn1 & mn2 \\
        \thickhline
        \rule{0pt}{1.1em}ALM & \textbf{2.35} & \textbf{3.20} & 2.48 & 3.35 & \textbf{2.35} & \textbf{3.20} \\
        ELS & 1.06 & 1.48 &0.95&1.32& \textbf{0.94} & \textbf{1.27} \\
        GDM  & \textbf{2.43} & \textbf{2.44} & 2.73 & 2.77 &2.45 & 2.47 \\
        MDR  & \textbf{4.42} & \textbf{5.71} & 4.82 & 6.02& \textbf{4.42}&\textbf{5.71} \\
        MND  & \textbf{0.82} & \textbf{1.01} & 0.94 & 1.11 &\textbf{0.82}&\textbf{1.01} \\
        NTD  & \textbf{1.27} &  1.47 & 1.47 & 1.75 &1.28 & \textbf{1.41} \\
        NYC  & \textbf{1.03} & \textbf{1.20} & 1.23 & 1.44 & \textbf{1.03}&\textbf{1.20} \\
        PDP & \textbf{2.18} & \textbf{2.92} & 2.57 & 3.38 &\textbf{2.18}&\textbf{2.92} \\
        PIC & \textbf{1.58} & 1.75 & 2.43 & 2.73 &1.62& \textbf{1.72} \\
        ROF & \textbf{2.18} & \textbf{2.28} & 2.92 & 3.09 &3.53&3.83 \\
        TOL & 1.15 & 1.20 & 1.98 & 2.04 &\textbf{1.10}& \textbf{1.14} \\
        TFG & \textbf{2.76} & \textbf{2.78} & 3.07 & 3.11 &3.53 & 3.59 \\
        USQ & 3.26 & 3.46 & 4.32 & 4.41 &\textbf{3.08}& \textbf{3.21} \\
        VNC & \textbf{4.96} & \textbf{7.31} & 5.70 & 7.93 &5.09 &7.48 \\
        YKM & 1.66 & 1.74 & 1.85 & 1.93 &\textbf{1.59}&\textbf{1.66} \\
        \thickhline
        \multicolumn{7}{c}{\rule{0pt}{1em}Baseline: RA \cite{chatterjee2018robust} + ROBA (100 iterations), }\\
        \multicolumn{7}{c}{mn1/2: mean angular error (deg) after $L_1$/$L_2$ alignment}\\
        \thickhline
    \end{tabular}
    \vspace{-0.5em}
    \end{center}
    \caption{
    Ablation study I:
    Undoing the square-rooting in \eqref{eq:our_cost2} (line \ref{line:square_rooting} of Alg. \ref{al:edge_cost}) worsens the accuracy for all datasets except ELS.
    Also, disabling the change of the step size $\alpha$ (line \ref{line:opti2} of Alg. \ref{al:proposed}) significantly worsens the accuracy for ROF and TFG.
    }
    \label{tab:ablation1}
\end{table}

\section{Discussions}
\label{sec:discussions}
\subsection{On error metrics}
As shown in Table \ref{tab:real_result}, the different alignment methods in \eqref{eq:alignment} result in different error values.
For example, the mean errors are always lower with the $L_1$ alignment because it yields the theoretically minimal mean error among all possible alignments. 
Also, the $L_1$ alignment often gives very small median errors because using both median and the $L_1$ alignment significantly diminishes the influence of large errors.
This was also reported in \cite{chatterjee2018robust}.
In Section \ref{sec:results}, we used the mean error after the $L_1$ alignment as our primary metric due to its moderate sensitivity to large errors.

\begin{table}[t]
\small
\begin{center}
    \begin{tabular}{l|ccc|ccc}
        \thickhline
        \rule{0pt}{1em} & \multicolumn{3}{c|}{\multirow{2}{*}{Baseline}} & \multicolumn{3}{c}{Without } \\
        & \multicolumn{3}{c|}{} & \multicolumn{3}{c}{approximating $\mathbf{g}$}\\
        Datasets&  mn1 & mn2 & Opti & mn1 & mn2 & Opti \\
        \thickhline
        \rule{0pt}{1.1em}ALM & 2.35 & 3.20 & \textbf{216} & \textbf{2.23} & \textbf{3.09} & 389 \\
        ELS & 1.06 & 1.48 &\textbf{42}&\textbf{0.97}& \textbf{1.32} & 76 \\
        GDM  & 2.43 & 2.44 & \textbf{76} & \textbf{2.36} & \textbf{2.40} & 135 \\
        MDR  & 4.42 & 5.71 & \textbf{52} & \textbf{4.21}& \textbf{5.41}& 93 \\
        MND  & 0.82 & 1.01 & \textbf{114} & \textbf{0.76} &\textbf{0.97}& 207 \\
        NTD  & 1.27 &  1.47 & \textbf{217} & \textbf{1.14} & \textbf{1.33} & 384 \\
        NYC  & 1.03 & \textbf{1.20} & \textbf{42} & \textbf{1.00} & 1.21& 75 \\
        PDP & 2.18 & 2.92 & \textbf{54} & \textbf{1.99} &\textbf{2.67}& 96 \\
        PIC & 1.58 & 1.75 & \textbf{617} & \textbf{1.43} & \textbf{1.61}& 1103 \\
        ROF & 2.18 & 2.28 & \textbf{154} & \textbf{1.17} &\textbf{1.29}& 274 \\
        TOL & 1.15 & 1.20 & \textbf{53} & \textbf{1.10} &\textbf{1.13}& 94 \\
        TFG & 2.76 & 2.78 & \textbf{1488} & \textbf{2.54} & \textbf{2.55} & 2664 \\
        USQ & 3.26 & 3.46 & \textbf{54} & \textbf{2.68} &\textbf{2.95}& 95 \\
        VNC & 4.96 & 7.31 & \textbf{221} & \textbf{4.39} &\textbf{7.03} & 395 \\
        YKM & 1.66 & 1.74 & \textbf{61} &\textbf{1.56}&\textbf{1.64} & 109\\
        \thickhline
        \multicolumn{7}{c}{\rule{0pt}{1em}Baseline: RA \cite{chatterjee2018robust} + ROBA (100 iterations), }\\
        \multicolumn{7}{c}{mn1/2: mean angular error (deg) after $L_1$/$L_2$ alignment}\\
        \multicolumn{7}{c}{Opti: Optimization time (in s) (line \ref{line:opti1}--\ref{line:opti2} of Alg. \ref{al:proposed}).}\\
        \thickhline
    \end{tabular}
    \vspace{-0.5em}
    \end{center}
    \caption{
    Ablation study II:
    In Alg. \ref{al:cost_and_gradient}, computing $\mathbf{g}$ without using the approximation \eqref{eq:gradient_assumption} improves the accuracy slightly.
    Exceptions are ROF and USQ where we observe large gains.
    In all cases, however, this significantly increases the optimization time.
    }
    \label{tab:ablation2}
\end{table}

\subsection{On convergence}
\label{subsec:convergence}
Most of the time, ROBA converges after 30--40 iterations, but it is subject to local minima.
Generally, the better the initial rotations, the better the final result. 
Since rotation averaging depends entirely on the relative rotation estimates, these input rotations must be accurate enough to achieve good results.
Empirically, we found that it is much better to have noisier input with fewer outliers than the other way around.
We also observed that sometimes a relatively small change in the total cost \eqref{eq:our_cost2} induces a non-negligible change in the rotational accuracy (see Fig. \ref{fig:evolution} and the supplementary material). 
This is somewhat in line with the observation in \cite{briales2018certifiably} for the two-view case.

\subsection{On robustness to outliers}
In this work, we did not consider outliers in the input, \ie, the image measurements and the initial rotation estimates.
We assumed that the outliers in the former have already been dealt with by a robust pose estimator and the latter by robust rotation averaging.
Taking into account also the outliers in our optimization is left for future work.

\subsection{On speed and scalability}
As discussed in Section \ref{subsec:eval_method}, we fixed the number of iterations at 100.
In some of the datasets, we observe that ROBA converges in much fewer iterations (see the supplementary material), so implementing the stopping criteria would reduce the runtime. 
We also note that our code was not highly optimized.
Possibly, one could increase the speed by vectorizing line \ref{line:M} in Al. \ref{al:edge_cost} using SIMD instructions.

As can be seen from Table \ref{tab:real_result}, the complexity of ROBA is linear in the number of edges.
To enhance the scalability, one could partition the view-graph and perform the local and the global optimization in parallel, as in \cite{{cui2017hsfm,zhu2018very}}.
This is left for future work.

\vspace{-0.2em}
\section{Conclusion}
\label{sec:conclusion}
\vspace{-0.2em}
In this work, we presented rotation-only bundle adjustment, a novel method for estimating the global rotations of multiple views independently of the translations and the scene structure.
We formulate the optimization problem by extending the two-view rotation-only method of \cite{kneip_rotation_ICCV} and solve it using the Adam optimizer \cite{adam}. 
As we decouple the rotation estimation from the translation and structure estimation, it is completely immune to their inaccuracies.
Our evaluation shows that (1) our method is robust to challenging configurations such as pure rotations and planar scenes, and (2) it consistently and significantly improves the accuracy when used after rotation averaging.

\clearpage
\newpage
{\small
\balance

}

\end{document}